\documentclass[letterpaper, 10 pt, conference]{ieeeconf}  

\IEEEoverridecommandlockouts                              

	\usepackage{cite}
	\usepackage{tikz}
	\usetikzlibrary{shapes.geometric, arrows}
	
	\usepackage{amsmath}

\newcommand{\fig}[1]{Fig.~\ref{fig:#1}}
\newcommand{\eqn}[1]{(\ref{eqn:#1})}
\title{\LARGE \bf
Deformation Control of a Deformable Object Based on Visual and Tactile Feedback
}

\author{Yuhao Guo$^{1}$ , Xin Jiang$^{1}$,~\IEEEmembership{member,~IEEE} and Yunhui~Liu$^{2}$,~\IEEEmembership{Fellow,~IEEE}
	\thanks{$^{1}$Yuhao Guo,  Xin Jiang are with Department of Mechanical Engineering and Automation, Harbin Institute of Technology, Shenzhen, HIT Campus Shenzhen 
University Town, Xili, Shenzhen, 518055, China.}
	\thanks{$^{2}$Yunhui Liu is with the Department of Mechanical and Automation Engineering, 
The Chinese University of Hong Kong, Hong Kong, China.}%
     \thanks{$^{*}$The corresponding author: Xin Jiang(x.jiang@ieee.org)}
}

	\usepackage{cite}
	\usepackage{tikz}
	\usetikzlibrary{shapes.geometric, arrows}
	\usepackage{amsmath}

\begin{document}
\maketitle
\thispagestyle{empty}
\pagestyle{empty}
\bibliographystyle{unsrt}
\begin{abstract}
In this paper, we presented a new method for deformation control of deformable objects, which utilizes both visual and tactile feedback. At present, manipulation of deformable objects is basically 
formulated by assuming positional constraints. But in fact, in many situations manipulation has to be performed under actively applied force constraints. This scenario is considered in this 
research. In the proposed scheme a tactile feedback is integrated to ensure a stable contact between the robot end-effector and the soft object to be manipulated. The controlled contact 
force is also utilized to regulate the deformation of the soft object with its shape measured by a vision sensor. The effectiveness of the proposed method is demonstrated by a book page 
turning and shaping experiment.
\end{abstract}
	
\section{INTRODUCTION}
In most of existing research on deformable object manipulation, control is formulated by utilizing the relationship between positional constraints and the corresponding deformation\cite{hou_review_2019}.
In these research, it is assumed that part of the soft object is pinned at robot end-effector. To satisfy this assumption, this part of the object is generally gripped tightly or physically pinned at the end-effector
such as the ways adopted in \cite{zhong_dual-arm_2019,hu_3-d_2019,hu_three-dimensional_2018,jin_robust_2019,alambeigi_autonomous_2019}. It may result to excessive force on the object. 
	
In many applications, soft objects are not allowed to be constrained with large local strain. For example, in many surgical operations, when using traditional surgical forceps, excessive gripping force exerted 
on soft tissues may lead to tissue damage as commented in\cite{guo_design_2017}. For many deformable objects, it is not easy to clamp them since their complex geometric shape or extreme sensibility to excessive 
local strain like the case for a balloon. In\cite{tsuchiya_pouring_2019}, grasping force control is proposed for holding a deformable container with flowable contents within it. Stable grasping is realized in the 
process when the contents is poured out.
	
The limitations in manipulation utilizing positional constraints can be summarized as follows: 1) Excessive force generated by clamping will cause damage to soft objects as demonstrated in \cite{guo_design_2017};
 2) For some soft objects, clamping is difficult to realize; 3) Like discussed in \cite{tsuchiya_pouring_2019}, some deformable objects (soft container) can not be described by a consistent model if active 
position constraints are assumed.   

A deformable object manipulation scheme realized by assuming force constraints may solve these problems, and it is suitable for more scenes where soft objects are manipulated. If the position constrained 
by active force is fixed, then it is equivalent to an active positional constraint.  A force constraint based approach can also tackle the situation where contact positions are not constant.
It is more difficult to manipulate a deformable object by active force constraints. The online estimation schemes widely used such as \cite{navarro-alarcon_automatic_2016}  will not be 
feasible since the contact point may change during manipulation. The estimation to the relationship between applied force and the corresponding deformation will be necessary as demonstrated 
in \cite{zaidi_interaction_2015}.

In this paper, we propose a scheme for deformation control which utilizes force constraints applied to a deformable object. This method is based on a similar assumption adopted in \cite{pestell_sense_2019}: 
large frictional force is achieved with a large contacting surface area. In addition  since the magnitude of friction force is affected by relative angle between fingertip and object, the finger tip is control to be always 
perpendicular to the contact surface in order to maximize the friction force. As a demonstration, book page turning and shaping will be realized using the proposed scheme. Different from the research in which 
page flipping is realized by utilizing a special mechanism or soft robot \cite{JiangSoftIROS2019}, in this research, it is realized by control the contact force.
	
The main contributions of this paper are: (1) We proposed a deformation control scheme for deformable objects, which is formulated by active force constraints. (2) We proposed a method to adaptively 
estimate the key tactile control parameters from repeated trials, which provides robustness to objects with unknown mechanical properties. (3) A demonstration of book page flipping was realized, which 
does not reply on any mechanisms but controlled contact condition between the finger and the object.  
	
The rest of the paper is organized as follows. In Section II, the related research is summarized. Section III contributes to describe a method for page turning.  The details of the shape control experiments 
are presented in Section IV. The paper ends with concluding remarks in section V.
\section{RELATED WORK}
Manipulation of rigid objects has made great achievement in recent years. In contrast, manipulation of deformable objects is still a challenge. Although manipulation of deformed objects based on force 
constraints is more challenging, many researchers have contributed a considerable number of references. The problems solved in these studies well demonstrate the aforementioned limitations
 in manipulation based on positional constraints.
	
Pestell et al. designed and built a set of tactile fingertips in\cite{pestell_sense_2019}, and present a simple linear-regression method for predicting roll and pitch of the finger-pad relative to a surface normal 
It also shows that the method can be applied to objects with unknown depths and shapes. And a grasp-control method with the self-made Modular Grasper is proposed.
	
In clinical and surgical settings, studies have been carried out in analyzing and measuring the forces exerted by these surgical instruments on tissues\cite{ponraj_soft_2018}. Histological evidence shows that 
excessive forces on tissues applied via these instruments may obstruct the blood supply and lead to localized necrosis, which affected the muscle function during and after the surgical procedure\cite{styf_effects_1998}\cite{taylor_impact_2002}. 
And currently, surgeons rely less on visual cues to determine the health of these retracted tissues\cite{fischer_ischemia_2006}. To deploy robots in surgery, Ponraj et al. developed a soft tactile sensor and mounted it on a retractor to maintain 
the stability of the applied force during operations \cite{ponraj_soft_2018}. Fischer et al. provided a means for sensory substitution with oxygenation sensors and force sensors incorporated on the working surfaces of surgical retractors and graspers in\cite{fischer_ischemia_2006}. These studies provide surgeons with sensory substitution and sensory augmentation during especially a robot assisted procedure.
	
There are many related studies in daily life scenarios. For instance, in \cite{geer_interaction_2020} the researcher used the tactile information provided by a single-point three-axis force sensor to recognize 
types of garments. It also helps to identify the whole manipulation process based on Dynamic Time Warping (DTW).  The contact  that occur at each time step is realized using a recurrent neural network (RNN). 
Delgado et al. realized stable, non-slip grasping based on tactile images and using servo control in\cite{delgado_tactile_2017}. The method they proposed analyzes and corrects deformations or changes in 
the objects due to robot arms' movements to control local deformations. Contact points are maintained by the tactile images which are obtained using a combination of dynamic Gaussians. This method can 
obtain a continuous representation of the tactile data, which makes the system more adaptable and suitable for tactile control. In\cite{tsuchiya_pouring_2019}, Tsuchiya et al. considered  manipulation of 
pouring flowable contents from deformable containers. Using two robotic hands with tactile information, they designed a grasping strategy. In this strategy two hands equipped with force sensors perform the 
grasping task together and the fingers of one hand apply the minimum force to suppress the deformation. With this method pouring was well realized.
	
\section{PAGE TURNING BASED ON TACTILE FEEDBACK}
In this research, tactile feedback based deformation control is demonstrated by manipulating a book page. In the subsequent 
sections, experiments including page turning and shaping are realized by a robot equipped with a tactile array sensor at its tip. 
	
\subsection{Experimental setup}
\begin{figure}[tb]
 \centering
 \includegraphics[width=\linewidth]{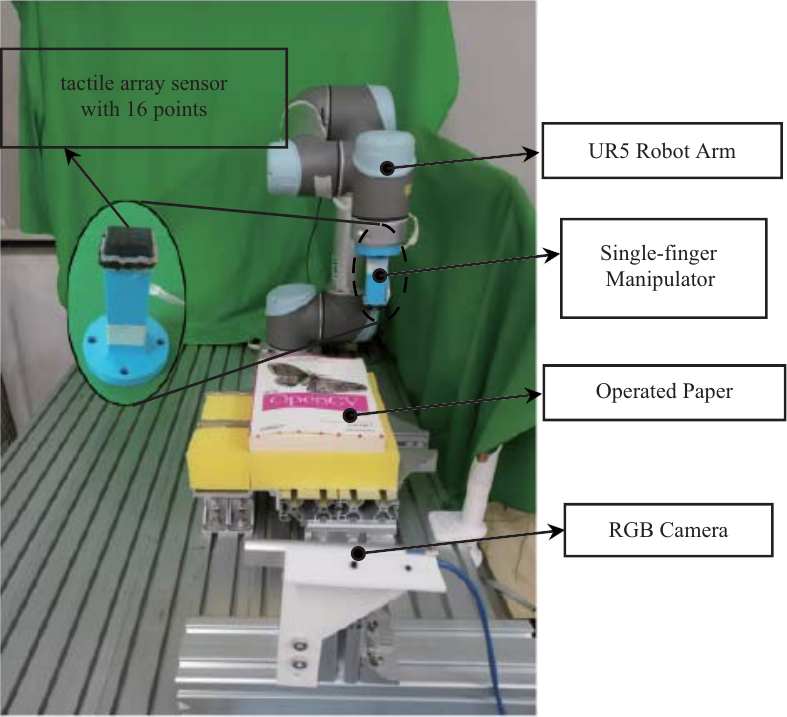}	
 \caption{The experimental setup composed of a UR5, a 16-points tactile array sensor, a RealSense Camera.}
\label{fig3.0}
\end{figure}
The experimental setup are composed of a robotic platform, a XELA uSkin tactile array sensor and a RealSense camera, as shown in Fig.\ref{fig3.0}. 
A UR5 manipulator is used as the robotic platform. The XELA uSkin tactile sensor is mounted at the end of a stick extended from the end-effector. 
In the subsequent part of the paper, we will call it a finger. In the tactile array sensor 16 three-axis force sensors are configured as 4x4 array. 
The sensor surface is covered by a rubber layer. A RGB-D sensor RealSense is used in the subsequent experiments to monitoring the shape variation of the book page.
The software of the whole experimental system is built based on ROS.
\subsection{Problem formulation}
In this research, page turning is expected to be realized by the end effector (we call it finger) in Fig. \ref{fig3.0}, that is to rub the page by friction force like a human does with her finger.
After the page is flipped, the robot is also expected to bend the paper to a desired shape without losing contact. From an analysis to the entire process, we know that in order to  
achieve the process the  finger should always be perpendicular to the paper at the contact point. When the finger moves, 
the friction and press force applied by the fingertip on the object provide constraints for deforming the object. 
	
\subsubsection{Realization of Vertical Pressing}
According to the contact force data obtained from the tactile sensor, a PID control is adopted to ensure that the finger is always perpendicular to the contact surface.
The force data from the tactile sensor can be modeled as two components: normal force and tangential force:
\begin{align}
		\mathbf{P_{f}} =\underbrace{\left[ \begin{matrix}
				p_{1} & p_{5}\\
				p_{2} & p_{6}\\
				p_{3} & p_{7}\\
				p_{4} & p_{8}
		\end{matrix}\right.}_{left}\underbrace{\left.\begin{matrix}
				p_{9} & p_{13}\\
				p_{10} & p_{14}\\
				p_{11} & p_{15}\\
				p_{12} & p_{16}
		\end{matrix}\right]}_{right} ,\ \mathbf{T_{f}} =\begin{bmatrix}
			t_{1} & \cdots  & t_{13}\\
			\vdots  & \ddots  & \vdots \\
			t_{4} & \cdots  & t_{16}
		\end{bmatrix}
\label{eqn:force_matrix}
\end{align}
where $\mathbf{P_f}$ is the press force data matrix and we call it pressure image, and $\mathbf{T_f}$ is the tangential force data matrix. We divide the pressure image 
into left and right regions as shown in \eqn{force_matrix} and sum them separately:
\begin{align}	
	 P_s^L=\sum_{i=1}^{8} p_i \,,\,  P_s^R=\sum_{i=9}^{16}p_i \,,\,P_s=\sum_{i=1}^{16}p_i \,,\, T_s=\sum_{i=1}^{16}t_i  
\end{align}	
where $P_s^L$ and $P_s^R$ represent the press force of the left and right region respectively, $P_s$ is the total press force exerted by the finger on the object, and $T_s$ is the 
total tangential force, which represents the friction between the finger and the object which is manipulated. 
	
Based on a PID algorithm and the robot servo system, the difference in press force ($ P_{dif}=P_s^L-P_s^R $) between the left and right regions is kept to 0, and the vertical 
pressing can be realized. As shown in \fig{fig3.2}, it can be assumed that the orientation of the finger directly affects the balance of the press force distribution. Thus, in 
order to balance the pressure distribution ($P_{dif}$ to 0), the angular velocity around X-axis is regulated following a PID control law formulated as : 
	\begin{figure}[tb]
		\centering
		\includegraphics[]{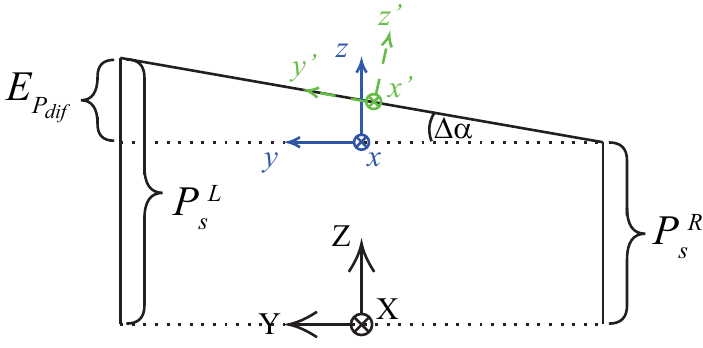}
		\caption{The diagram of  realizing of vertical pressing. The coordinate system XYZ in the figure represents the global coordinate system where the finger is located. The coordinate system $xyz$ and $x'y'z'$ represent the change of the finger's orientation before and after a servo cycle.}
		\label{fig:fig3.2}
	\end{figure}
\begin{align}
		E_{P_{dif}}& = P_{dif}  - P_{dif}^{des} = P_{dif} - 0 = P_{dif} \\
		& E_{P_{dif}}\xrightarrow{PID} \omega _{x} \ ,
		\ \Delta \alpha =\Delta t\times \omega _{x}
\label{eqn:eqn3}
\end{align}
where $\omega_{x}$ refers to the angular velocity command around the X-axis, $\Delta \alpha$ refers to rotation of the finger around the X-axis in each servo cycle, and $\Delta t$ is the servo period. 
\subsubsection{Movement of the finger}
Besides adjusting its orientation, the finger also needs to move along the normal direction of contact surface in order to adjust the press force to the desired value. In addition, the finger 
also needs to perform a tangential movement to add a tangential constraint force according to the requirement for deformation control. We refer to these two types of motion as 
\textit{vertical motion} and \textit{tangential motion}, as shown in \fig{fig3.3}. The vertical motion depends on the press force ($P_s$ in (2)), while the tangential motion is determined 
by a constant velocity, as in: 
	\begin{figure}[t]
		\centering
		\includegraphics[]{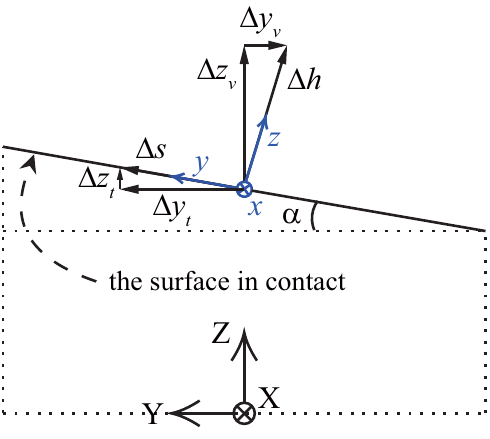}		
		\caption{The diagram of tangential motion and vertical motion. The coordinate system XYZ in the figure represents the global coordinate system where the finger is located. The coordinate system $xyz$ represent the finger's orientation.}
		\label{fig:fig3.3}
	\end{figure}
\begin{align}
		E_{P_{s}} =P_{s} - P^{des}_{s}& \ ,\ E_{P_{s}}\xrightarrow{PID} \mathbf{V_v} \\
		\Delta \mathbf{h}=\Delta t \times \mathbf{V_v}\ &,\ 
		\Delta \mathbf{s}=\Delta t \times \mathbf{V_{t}}
\end{align}
where $P^{des}_{s}$ is the referece of $P_s$, $E_{P_{s}}$ is the error between $P_s$ and $P^{des}_{s}$, $\mathbf{V_v}$ and $\mathbf{V_t}$ are the velocities of vertical and tangential motion, and $\Delta \mathbf{h}$ and $\Delta \mathbf{s}$ are the moving distance along the vertical and tangential direction in a servo period ($\Delta t$).
The motion controller of the end effector is shown in \fig{fig3.4}. 

	\begin{figure}[tb]
		\centering
		\includegraphics[width=\linewidth]{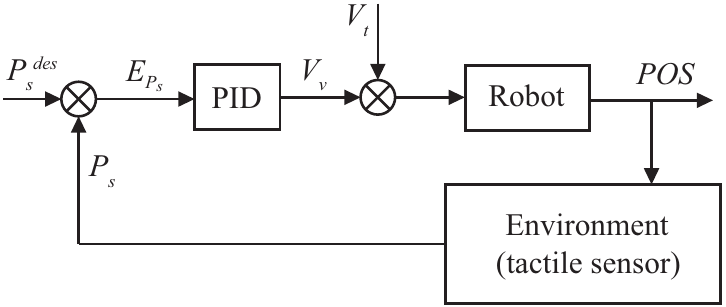}
		\caption{Block diagram of the motion controller. $POS$ is the position of the finger.}
		\label{fig:fig3.4}
	\end{figure}
\subsection{The strategy for page turning}
	\begin{figure*}[tb]
		\centering
		\includegraphics[width=\linewidth]{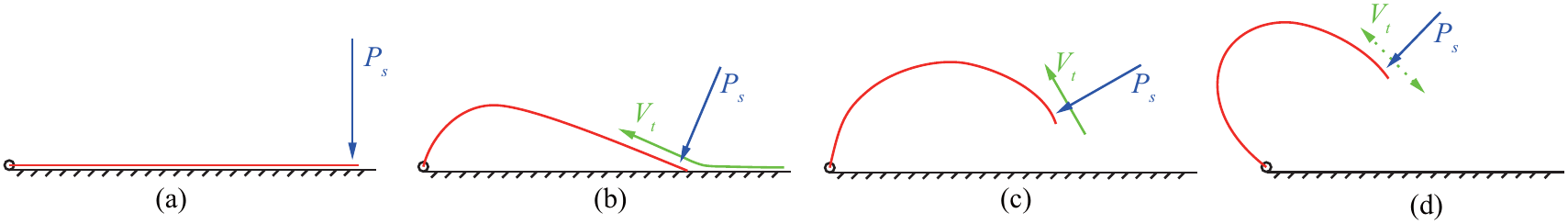}
		\caption{The four basic stages of paper manipulation: (a) \textit{\textbf{press}} the paper, (b) \textit{\textbf{rub}} the paper, (c) turn \textit{\textbf{up}} the paper, (d) \textbf{\textit{shape control}}. $P_s$ is the press force exerted by the finger on the paper, as shown in (2), and $V_t$ is the velocity of tangential motion as shown in (6). The lines in black represents the support on which the paper is placed.}
		\label{fig:fig4.0}
	\end{figure*}
\begin{figure*}[tb]
		\centering
		\includegraphics[width=\linewidth]{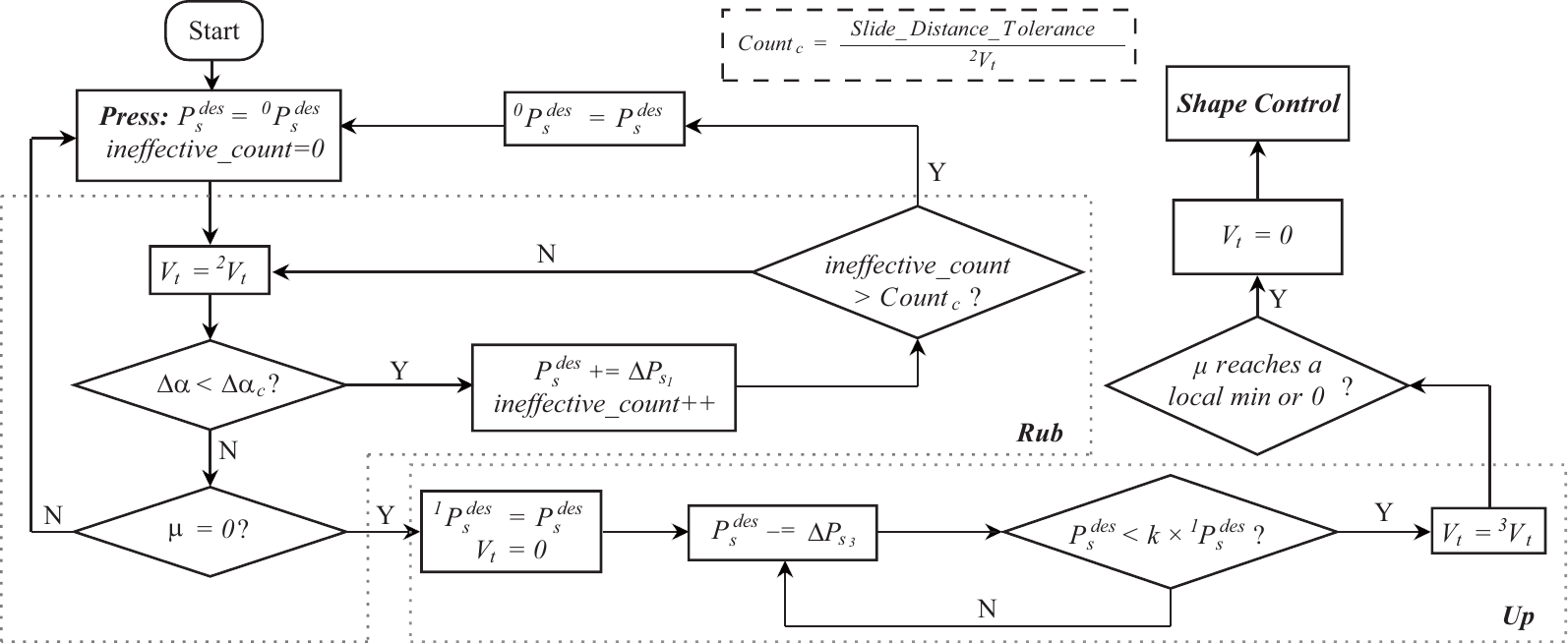}
		\caption{Block diagram of our proposed algorithm includes four manipulation stages as shown in \fig{fig4.0}: \textbf{\textit{Press}}, \textbf{\textit{Rub}}, \textbf{\textit{Up}}, \textbf{\textit{Shape Control}}.}
		\label{fig:fig4.1}
\end{figure*}

According to an analysis to similar manipulation process completed by human, we divide the entire process of page turning into four sequential stages as shown in \fig{fig4.0}. In the experiment, the shape of book page is described by its cross-sectional shape 
as illustrated in \fig{fig4.0}. The transition between successive stages are determined by the coefficient of friction $\mu (=T_{s} /P_{s})$.
\subsubsection{\textbf{Press}}
In the first stage, the finger is moved in the vertical direction to make the normal press force to be the desired one $^{1} P^{des}_{s}$. The tangential velocity $\mathbf{V_t}$ in this process is set to 0. 
The reference press force $^{1} P^{des}_{s}$ plays an important role in the next stage. The success of the next stage depends on whether the value of $^{1} P^{des}_{s}$ is reasonable. In order to 
improve the robustness, we also proposed an algorithm to adaptively determine the value of $^{1} P^{des}_{s}$, which will be introduced later in this section.
	
\subsubsection{\textbf{Rub}}
In the second stage, a constant tangential velocity $ ^{2}\mathbf{V_{t}}$ is specified. The finger is moved tangentially with the specified velocity while the pressure is maintained at $^{1} P^{des}_{s}$. The tangential 
motion will stop until the termination condition $\mu=0$ is satisfied. 
	
\subsubsection{\textbf{Up}}
In third stage, the finger flips up the paper to make it completely separate from the support. The manipulation strategy adopted here is to reduce the pressure $P_s$ from $^{1} P^{des}_{s}$ to $^{3} P^{des}_{s}$ 
and then perform tangential motion until the termination condition of this stage is satisfied. Similar to $^{1} P^{des}_{s}$, $^{3} P^{des}_{s}$ is also determined adaptively by our algorithm. 
And the termination condition of this stage is that $\mu$ reaches its local minimum or 0.
	
\subsubsection{\textbf{Shape Control}}
Through the first three stages, the paper has been completely turned up. The next object is to verify the whether a shape control can be realized by the 
force constraints exerted. Thus the robot will try to deform the book page to a pre-defined shape. 
\begin{figure*}[tb]
		\centering
		\includegraphics[width=\linewidth]{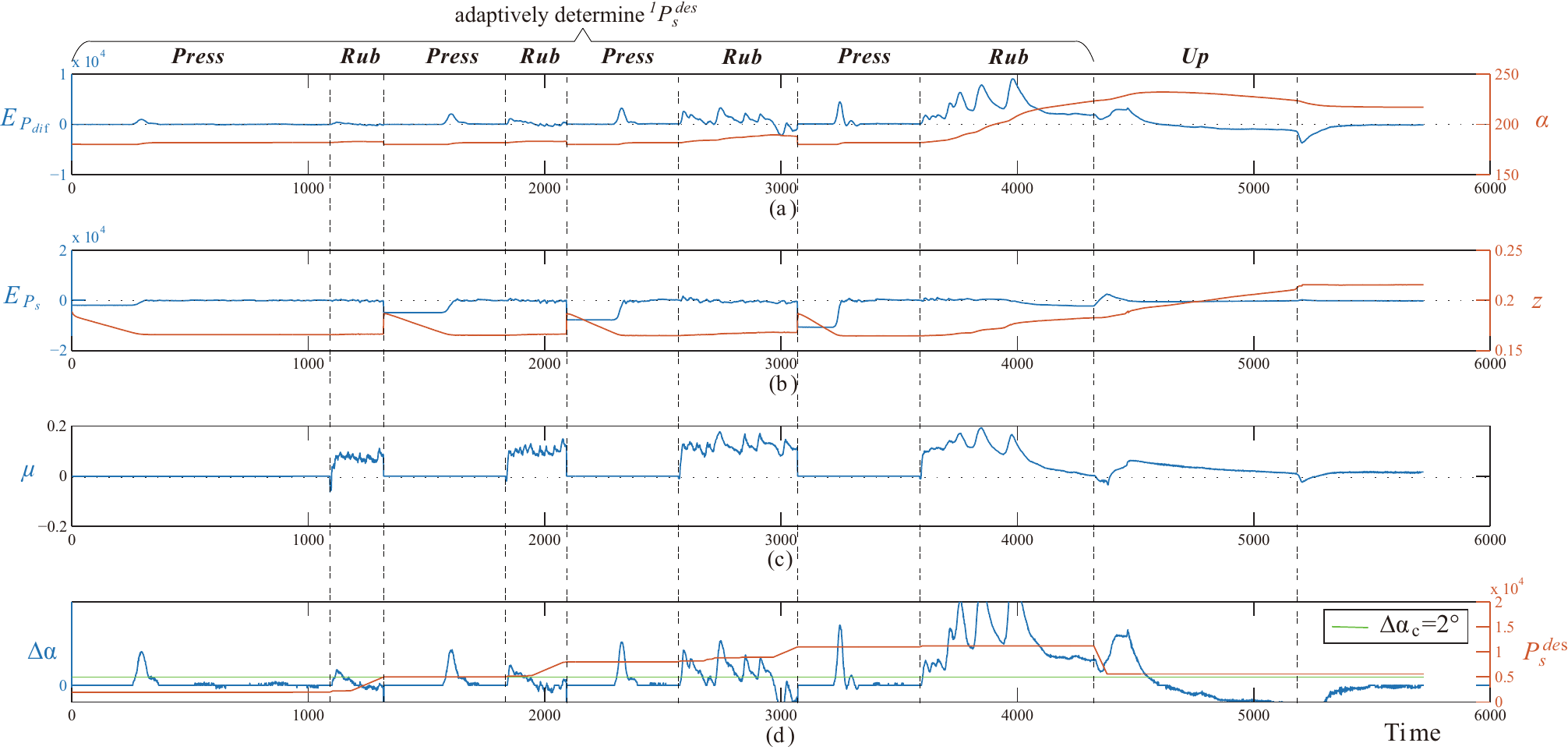}
		\caption{The data of paper manipulation experiment containing the first three stages. Adaptive determination of $^1P_s^{des}$ completes in stage Press ans Rub, and $^3P_s^{des}$ is determined between stage Rub and Up. (a) $E_{P_{dif}}$ and $\alpha$ as in (4) realize vertical pressing. (b) $E_{P_s}$ as in (5) determines the vertical motion, and $z$ indicate the finger's movement. (c) $\mu$ is used as transition conditions between different stages. (d) The adaptive determination of $^1P_s^{des}$ and $^3P_S^{des}$ using $\Delta \alpha$.}
		\label{fig:fig4.2}
\end{figure*}
	
\subsection{Adaptive force determination algorithm}
The reference pressure $^{1} P^{des}_{s}$ in the first stage and $^{3} P^{des}_{s}$ in the third stage are determined adaptively by the algorithm. 
The optimal coefficients of $^{1} P^{des}_{s}$ and $^{3} P^{des}_{s}$ all should be determined according to the surface roughness and the deformation properties of the paper, such as stiffness. 
Since no prior information on the object are assumed,  $^{1} P^{des}_{s}$ and $^{3} P^{des}_{s}$ are determined adaptively.
As for $^{1} P^{des}_{s}$, it needs to be large enough to provide necessary flip force to rub the paper. On the other hand, an excessive force setting for $^{1} P^{des}_{s}$ may lead the robot to fail in separating 
only one page or even in completing stage 2. We realized the algorithm by introducing two concepts: \textit{Effective-rubbing} and \textit{Ineffective-rubbing count}, as follows:
\begin{itemize}
\item \textit{Effective-rubbing}: At first, make the finger to press on the paper with a small force ($P_s^{des}=$$^{0}P_s^{des}$), as described in stage 1. And then, during the rubbing action in stage 2, 
we monitor the effect of rubbing in real time. When a rubbing action effectively deform the book page, it is called \textit{effective-rubbing}. Otherwise, it is called \textit{ineffective-rubbing} if the finger
 just slides on the surface of the paper. When an ineffective-rubbing occurs, pressure will increase ($P_s^{des}  += \Delta P_s$). The effective-rubbing and ineffective-rubbing are distinguished by 
using the $\Delta \alpha$ in \eqn{eqn3}, because the deformation of paper will cause a change in the orientation of the finger. We determined the critical value of $\Delta \alpha _c$  experimentally. 
The experiments indicated that the best effect to distinguish two kinds of rubbing is achieved $\Delta \alpha _c=2^\circ$. 
		
\item \textit{Ineffective-rubbing counter}: Ineffective-rubbing means sliding on the paper during a certain servo cycle, and the pressure will increase ($P_s^{des}  += \Delta P_s$) when it occurs.
Too many times of ineffective-rubbing will cause the sliding distance to exceed the tolerance. It will lead the page turning manipulation fail. 
Therefore, we introduce an ineffective-rubbing counter to mark the total sliding distance in one trial of "rub". It is expressed as the relative distance between the finger and the starting position of the rubbing action. 
If the sliding distance reaches the threshold, it means that even if the movement continues, there is no possibility of effectively completing the rubbing action. 
The manipulation will restart from stage 1 and the initial pressure ($^{0}P_s^{des}$) will be updated as the latest pressure ($P_s^{des}$). 
\end{itemize}

At the end of the stage 2, $^{1} P^{des}_{s}$ is determined as $P_s^{des}$ at this time. As for the adaptive determination of $^{3} P^{des}_{s}$, since they all depend on the stiffness and 
roughness of the paper, $^{3} P^{des}_{s}$ and $^{1} P^{des}_{s}$ are assumed to be positively correlated ($^{3} P^{des}_{s} \propto $$^{1} P^{des}_{s}$). So we simply defined their relationship as 
$^{3} P^{des}_{s} = k \times$$^{1} P^{des}_{s}$. The parameter is determined as $k=0.5$ experimentally. The block diagram of our proposed algorithm is shown in \fig{fig4.1}.
Many experiments have been done using the algorithm, as recorded in Table \ref{tab:table1}, and the results obtained show that the algorithm is effective. 

\begin{figure}[tb]
	\centering
	\includegraphics[width=0.9\linewidth]{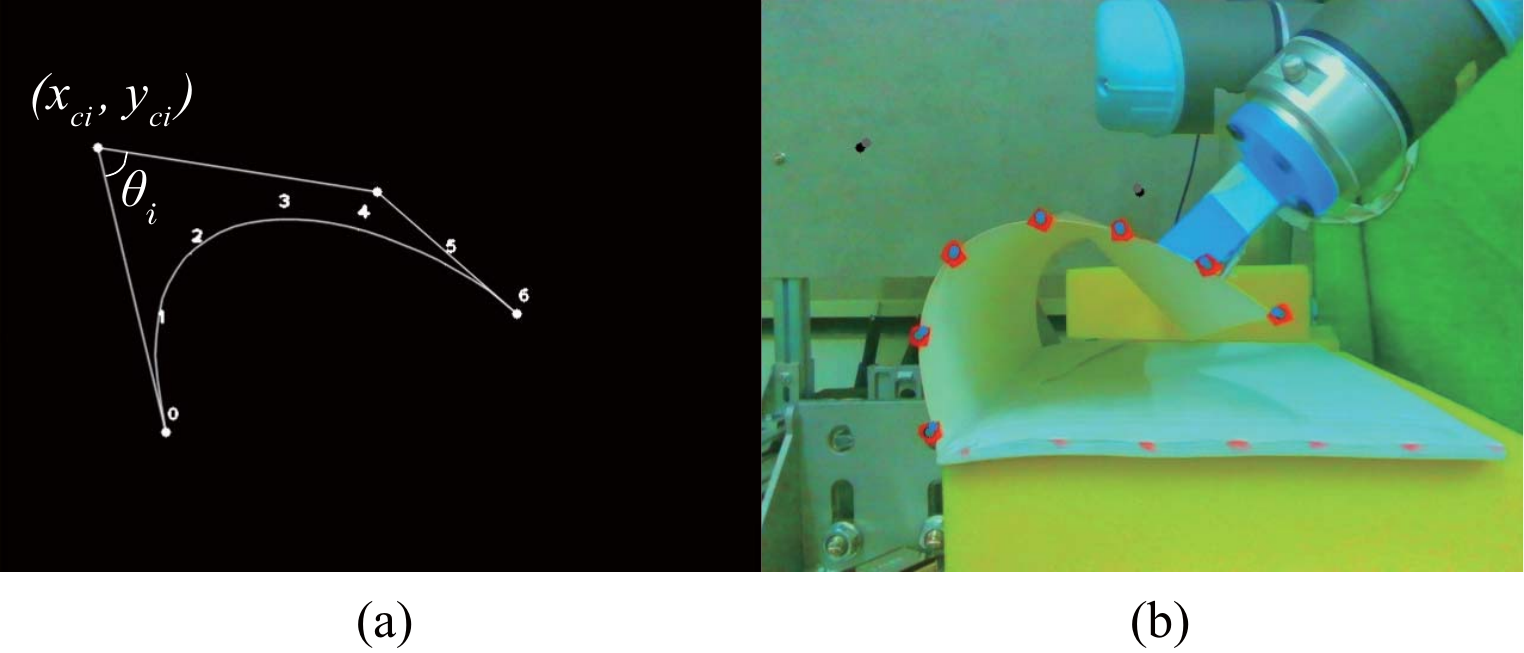}
	\caption{The result of parameterization includes fitted curve and its control points. $\theta_{i}$ is the interior angle between the lines of control points. $(x_{ci},y_{ci})$ is the coordinate of control point $i$. The number 0-6 mark the location of the data point 0 to 6.}
	\label{fig:fig5.0}
\end{figure}

\begin{table}[h]
	\centering
	\begin{tabular}{c|c|c}
		\hline
		\begin{tabular}[c]{@{}c@{}}\textbf{Experimental}\\ \textbf{samples}\end{tabular} & \begin{tabular}[c]{@{}c@{}}\textbf{Critical pressure for}\\\textbf{rubbing up the paper}\end{tabular} & \begin{tabular}[c]{@{}c@{}}\textbf{Stiffness}\end{tabular} \\ \hline
		Book 1                                                         & $^{1} P^{des}_{s} = 2180$                                             & low                                                          \\ \hline
		Book 2                                                         & $^{1} P^{des}_{s} = 2160$                                             & low                                                          \\ \hline
		Book 3                                                         & $^{1} P^{des}_{s} = 5200$                                             & medium                                                       \\ \hline
		Book 4                                                         & $^{1} P^{des}_{s} = 5440$                                             & medium                                                       \\ \hline
		Book 5                                                         & $^{1} P^{des}_{s} = 8240$                                             & high                                                         \\ \hline
		Book 6                                                         & $^{1} P^{des}_{s} = 8200$                                             & high                                                         \\ \hline
		Book 7                                                         & $^{1} P^{des}_{s} = 9500$                                             & high                                                         \\ \hline
	\end{tabular}
	\caption{Page turning experiments record.}
	\label{tab:table1}
\end{table}

One of the experiments is shown in \fig{fig4.2}. As indicated in the figure, the optimal reference of the press force $P^{des}_{s} $ is determined after four trials of "press then rub".
The previous three trials are terminated separately when the ineffective-rubbing counter indicates reaching to the threshold of counter. It implies that the tip has slidden too far on the paper. In the rub phase of the fourth trial, it can be confirmed that with the latest reference of $ P^{des}_{s}$, the tangential motion has started 
to deform the paper effectively (indicated by the variation of $\Delta \alpha$). This phenomenon continues until the  friction coefficient $\mu$ becomes to be zero, which indicates that the paper is  separated from the support.
The procedure proceeds to the next stage of "up" and the reference press force reduces by half. With the determined value, the paper is successfully rubbed up.


\begin{figure*}[tb]
	\centering
	\includegraphics[width=\linewidth]{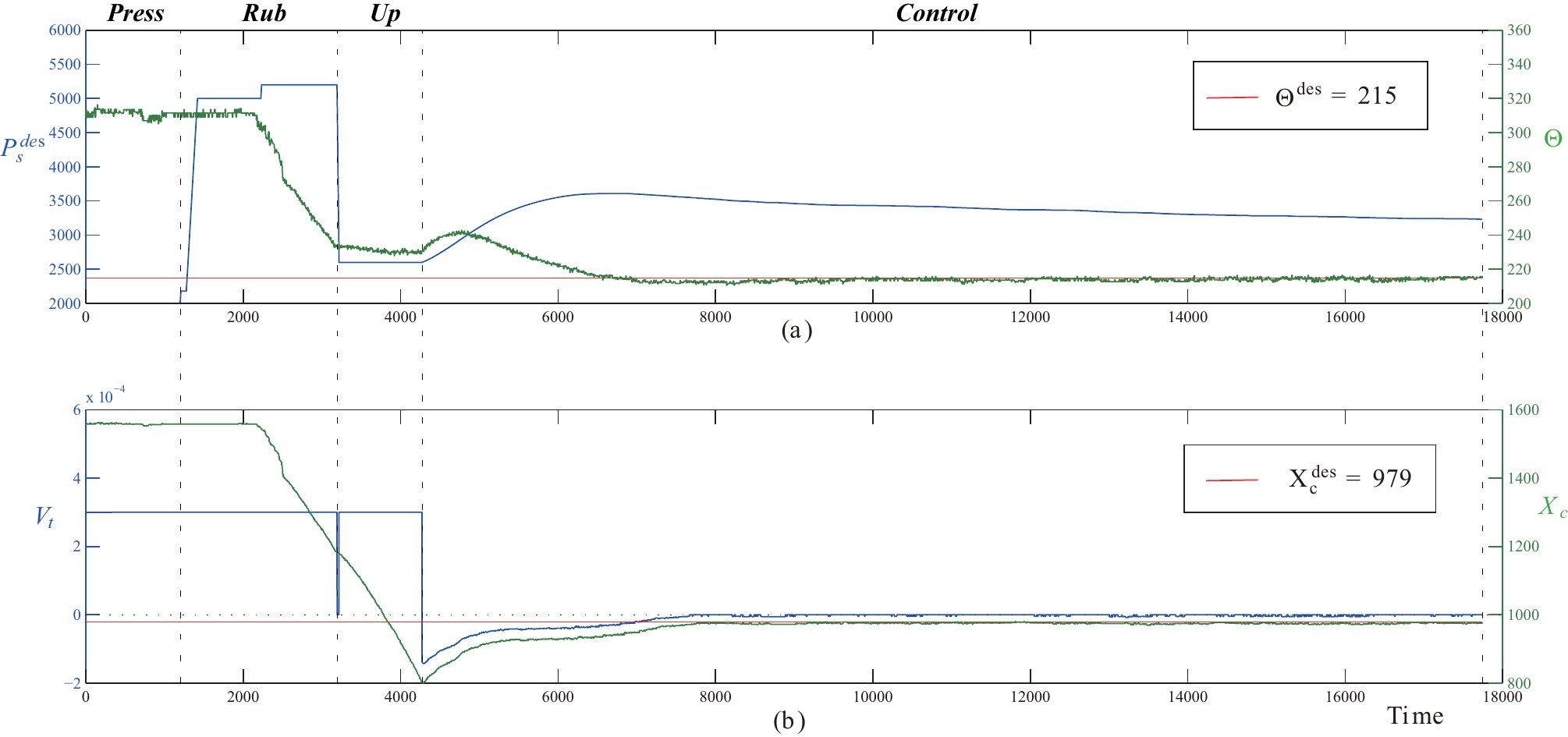}
	\caption{Bending the paper to a pre-determined shape: $(\Theta ^{des},X_c^{des})=(215,979)$. }
	\label{fig:fig5.1}
\end{figure*}

\section{DEFORMATION CONTROL}
The second part of this research is the deformation control based on force constraints. It is demonstrated  by bending the flipped paper to a desired shape.
The experiment is the fourth stage in the scenario described in \fig{fig4.0}. This deformation control method is model-independent and the RGB images from a 
Realsense camera is used as the feedback.

Shape of a paper is represented by its cross-sectional contour in our experiment,  which is measured by monitoring the markers attached on its edge.
We use B\'ezier curve to parameterize the measured shape. The parameterization process is completed by processing the RGB image with OpenCV. 
The result of parametric coefficients and the fitted curve are shown in \fig{fig5.0}.

The paper shape curve is uniquely represented by a numerical pair $(\Theta ,X_c)$. 
\begin{align}
	\Theta =\sum \theta _{i} \ ,\ X_c=\sum x_{ci} \\
	\ paper\ shape\Longleftrightarrow ( \Theta ,X_c)
       \label{eqn:eqn5}
\end{align}
where $\theta _{i}$ is the interior angle and $x_{ci}$ is the abscissa of the $i th$ control points in the image, as shown in \fig{fig5.0}. 
The two variable represent deformation status and the exact position of the deformed contour respectively.  We assume that 
the status of deformation $\Theta$ is uniquely determined by the pressure ($P_s^{des}$). And the position of the deformed contour $X_c$ with respect to global coordinate frame is determined by the position of the contact point, which can be adjusted by the tangential movement ($V_t$).  Thus control of the two 
variables is be solved by two independent controllers. The corresponding controllers are realized by PID simply, as in:
\begin{align}
	E_{\Theta } =\Theta -\Theta ^{des} \ ,\ E_{X_{c}} =X_{c} -X^{des}_{c}\\
	E_{\Theta }\xrightarrow{PID} \Delta {^{4}P^{des}_{s}} \ ,\ E_{X_{c}}\xrightarrow{PID} {^{4} V_{t}}
\label{eqn:eqn6}
\end{align}
where $\Delta{^{4} P^{des}_{s}}$ is the change of the reference press force in stage 4, and ${^{4} V_{t}}$ is the tangential velocity command.
In experiments, the actual press force can not converge to its specified value $P^{des}_{s}$ immediately. Thus the PID control for $\Theta$ is 
formulated to change the difference of reference between to successive sample interval. 
We have done several experiments successfully with different papers as shown in Table \ref{tab:table2}, and one of the experiment result as shown in \fig{fig5.1}. 
It indicates that the deformation control algorithm is effective.

\begin{table}[h]
	\centering
	\begin{tabular}{c|c|c|c}
		\hline
		\textbf{\begin{tabular}[c]{@{}c@{}}Experimental\\ samples\end{tabular}} & \textbf{\begin{tabular}[c]{@{}c@{}}Critical pressure for\\ rubbing up the paper\end{tabular}} & \textbf{\begin{tabular}[c]{@{}c@{}}Target\\ $(\Theta ^{des},X^{des}_{c})$\end{tabular}} & \textbf{\begin{tabular}[c]{@{}c@{}}Stiffness\end{tabular}} \\ \hline
		Book 1                                                                  & $^{1} P^{des}_{s} = 2380$                                                      & $(183.198,780)$                                                                         & low                                                                   \\ \hline
		Book 2                                                                  & $^{1} P^{des}_{s} = 5200$                                                      & $(215.0,979)$                                                                           & medium                                                                \\ \hline
	\end{tabular}
	\caption{Deformation control experiments record.}
	\label{tab:table2}
\end{table}

\section{CONCLUSION AND DISCUSSION}
In this paper, we presented a method to flip up a book page and control its deformation. The proposal is realized by using tactile feedback.
It demonstrates that the tasks including paper flipping and shaping can be realized by simultaneously controlling three variables: pressure distribution, 
net press force, tangential velocity of the finger tip. The three control objectives are regulated by rotating the finger around its x axis, moving the finger tip long the 
normal and tangential direction of the contact surface respectively. Since the motion commands generated following this rule do not interfere with each other.
Therefore the three control objectives are realized by deploying three independent PID controllers. All the key parameters used in these controls are determined 
adaptively through repeated trials. This property makes the proposal useful for manipulating objects in the situations where no prior knowledge is available.
	
 \addtolength{\textheight}{-9cm}   
\section*{ACKNOWLEDGMENT}
This work was supported by Natural Science Foundation of China (Grant No.61873072), National Natural Science Foundation of China (Grant No.U1713202).
\bibliography{reference}
\end{document}